\documentstyle[a4,11pt,lingmacros,avm]{article}
\newcommand{\avmsp}{\vspace*{1ex}}
\newcommand{\attr}{\sc}
\newcommand{\exm}{\it}
\avmsortfont{\small\it}
\avmvalfont{\rm}
\avmfont{\attr}

\begin{document}
\setcounter{footnote}{1}
\vspace*{-5mm}
\begin{center}
{\Large\bf A Categorial Framework for Composition in Multiple Linguistic
Domains}
\protect\footnotetext{This research is
supported in part by grants from
Scientific and Technical Research Council of Turkey (contract no. EEEAG-90),
NATO Science for Stability Programme (contract name TU-LANGUAGE), and
METU Graduate School of Applied Sciences.}\\ \vspace*{.5ex}
\vspace*{2mm}
{\bf H. Cem Boz\c{s}ahin} and {\bf Elvan G\"{o}\c{c}men}\\
Laboratory for the Computational Studies of Language\\
Computer Engineering Department\\
Middle East Technical University\\
06531 Ankara, Turkey\\ {\tt \{bozsahin,elvan\}@lcsl.metu.edu.tr}
\end{center}

\begin{abstract}
We describe a computational framework for a grammar architecture
in which different linguistic domains such as morphology, syntax,
and semantics are treated not as separate components but
compositional domains. Word and phrase formation are modeled as
uniform processes contributing to the derivation of the semantic form.
The morpheme, as well as the lexeme, has lexical representation
in the form of semantic content, tactical constraints, and phonological
realization. The model is based on Combinatory Categorial Grammars.
\end{abstract}
%\maketitle
%\bibliographystyle{acl}
\section{Introduction}
\label{intro}
The division of morphology and syntax in agglutinative languages is
difficult compared to relatively more isolating languages. For
instance, in Turkish, there is a significant amount of interaction between
morphology and syntax. Typical examples are: causative suffixes change the
valence of the verb, and the reciprocal suffix subcategorize the verb
for a noun phrase marked with the comitative case. Moreover, the head
that a bound morpheme modifies may be not its stem but a compound
head crossing over the word boundaries, e.g.,

\enumsentence{\label{childex}
\shortex{3}{{\exm iyi} & {\exm oku-mu\c{s}} & {\exm \c{c}ocuk}}
           {well & read-REL & child}
           {'well-educated child'}
}

In~(\ref{childex}), the relative suffix {\exm -mu\c{s}}
 (in past form of subject participle)
modifies [{\exm iyi oku}] to give the scope
[[[{\exm iyi oku}]{\exm mu\c{s}}] {\exm \c{c}ocuk}].
If syntactic composition is performed after morphological composition,
we would get compositions
such as [{\exm iyi} [{\exm okumu\c{s} \c{c}ocuk}]] or
[[{\exm iyi okumu\c{s}}] {\exm \c{c}ocuk}], which yield ill-formed
semantics for this utterance.

As pointed out by Oehrle~\cite{oehrle88,oehrle94}, there is no reason to assume
a layered grammatical architecture which has linguistic division of
labor into components acting on one domain at a time. As a
computational counterpart of this idea, rather than
treating morphology, syntax and semantics in a cascaded manner,
we integrate the process models of morphology and syntax, providing
semantic composition in parallel.
The model,
which is based on
Combinatory Categorial Grammars (CCG)~\cite{ades82,steedman85},
uses the morpheme as the building
block of composition at all three linguistic domains.

\section{Morpheme-based Compositions}
\label{morpheme}

When the morpheme is given the same status as the lexeme in terms
of its lexical, syntactic, and semantic contribution, the distinction
between the process models of morphotactics and syntax disappears.
In this case, new scoping problems arise in word and phrase formation.

CG accounts of scoping problems concentrate on syntactic and semantic
issues such as quantifier scoping~\cite{oehrle94,pereira90}. In word
formation, morphological bracketing paradoxes are introduced by
lexicalized composite affixes which require mixed
compositions~\cite{moortgart88}. However, the scoping problems in
morphosyntax go beyond bracketing paradoxes as they may also
produce different semantic forms.
Consider the example in~(\ref{npex}):

\enumsentence{\label{npex}
\shortexnt{3}{{\exm uzun} & {\exm kol-lu} & {\exm g\"{o}mlek}}
           {long & sleeve-ADJ & shirt}
}

\begin{figure}
\small
\begin{center}
\begin{tabular}{|lll|} \hline
lexical entry & syntactic category & semantic category \\ \hline
{\exm uzun} &  $n/n$ & $\lambda p.long(p(z))$ \\
{\exm kol} & $n$ & $\lambda x.sleeve(x)$   \\
{\exm -lu} & $(n/n)\setminus n$ & $\lambda q.\lambda r.r(y,has(q))$ \\
{\exm g\"{o}mlek} & $n$  & $\lambda w.shirt(w)$\\  \hline
\end{tabular}\vspace*{3mm}

(1a)\hspace*{1cm}\begin{tabular}{llll}
{\exm uzun} & {\exm kol} & {\exm -lu} & {\exm g\"{o}mlek} \\
\cline{1-2}
$n$ \\
\cline{1-3}
$n/n$ \\
\cline{1-4}
$n$\\
\end{tabular}
\vspace*{3mm}

$shirt(y,has(long(sleeve(z))))$ = 'a shirt with long sleeves'
\vspace*{6mm}

(1b)\hspace*{1cm}\begin{tabular}{llll}
{\exm uzun} & {\exm kol} & {\exm -lu} & {\exm g\"{o}mlek} \\
\cline{2-3}
& $n/n$ \\
\cline{2-4}
& $n$ \\
\cline{1-2}
$n$\\
\end{tabular}
\vspace*{3mm}

$long(shirt(y,has(sleeve(z))))$ = 'a long shirt with sleeves'
\end{center}
\caption{Scope ambiguity of a nominal bound morpheme}
\label{gomlekcompose}
\end{figure}

Two different compositions\footnote{derived and basic
categories in the examples
are in fact feature structures; see section~\ref{fssect}. We use
$\frac{x\ \ y}{z\ \ \ }$ to denote the combination of categories $x$ and $y$
giving the result $z$}
in CCG formalism are
given in Figure~\ref{gomlekcompose}. Both interpretations are
plausible, with (1a) being the most likely in the
absence of  a long pause after the first adjective. To account for
both cases, the suffix {\exm -lu} must be allowed to modify the head
it is attached to (e.g., 1b in Figure~\ref{gomlekcompose}),
or a compound head encompassing the word boundaries (e.g., 1a
in Figure~\ref{gomlekcompose}).

Example~(\ref{vpex}) shows a composition with a verbal head.
Figure~\ref{konustucompose} depicts the CCG treatment of this example.
The verb {\exm konu\c{s}} does not
subcategorize for a dative noun phrase (cf. example~\ref{vpex}b);
{\exm kad{\i}na} is the argument of {\exm d\"{o}n}.
In this case, the adverbial suffix {\exm -erek} must modify
[{\exm kad{\i}na d\"{o}n}]
to obtain the correct reading.
\eenumsentence{\label{vpex}
\item[a.]\shortex{3}{{\exm kad{\i}n-a} & {\exm d\"{o}n-erek} & {\exm
konu\c{s}-tu}}
                    {woman-DATIVE & turn-ADV & talk-TENSE}
                    {'Facing the lady, (he/she) talked.'}
\item[b.]* {\exm kad{\i}na konu\c{s}tu}
}

\begin{figure}
\small
\begin{center}

\begin{tabular}{|lll|} \hline
lexical entry & syntactic category & semantic category \\ \hline
{\exm kad{\i}n} & $n$  & $\lambda x_1.female(x_1)$\\
{\exm -a} & $(s\setminus n)/(s\setminus n)\setminus n$  &
 $\lambda p_3.\lambda p_2.p_2(to(p_3(m)))$\\
{\exm d\"{o}n} & $(s\setminus n)\setminus n$ &
 $\lambda x_4.\lambda x_5.turn(x_5,x_4)$\\
{\exm -erek} &
 $(s\setminus n)/(s\setminus n)\setminus((s\setminus n)\setminus n)$ &
 $\lambda p_6.\lambda p_7.p_7(\lambda x_6.(x_6,by(p_6(x_6))))$\\
{\exm konu\c{s}} & $s\setminus n$ & $\lambda x_8.speak(x_8)$\\
   {\exm -tu} & $(s\setminus n)\setminus (s\setminus n)$ &
$\lambda p_9.p_9(z,past)$ \\ \hline
\end{tabular}\vspace*{3mm}

\begin{tabular}{llllll}
{\exm kad{\i}n} & {\exm -a} & {\exm d\"{o}n} & {\exm -erek} & {\exm konu\c{s}}
&
   {\exm -tu}\\
\cline{1-2}
$(s\setminus n)/(s\setminus n)$ \\
\cline{1-3}
$(s\setminus n)\setminus n$ \\
\cline{1-4}
$(s\setminus n)/(s\setminus n)$ \\
\cline{1-5}
$s\setminus n$ \\
\cline{1-6}
$s\setminus n$ \\
\end{tabular}
\vspace*{3mm}

$\lambda x_6.speak(x_6,by(turn(x_6,to(female(m))))$ =
'facing the lady, (he/she) talked'
\end{center}
\caption{Composition with a verbal bound morpheme}
\label{konustucompose}
\end{figure}

\section{Multi-domain Combination Operator}

Oehrle~\cite{oehrle88} describes a model of multi-dimensional composition
in which every domain $D_i$ has an algebra with a finite set
of primitive operations $F_i$. As indicated by Turkish data in
sections~\ref{intro} and~\ref{morpheme}, $F_i$ may in fact have a
domain larger than---but compatible with---$D_i$.

In order to perform morphological and syntactic compositions in a
unified (monostratal) framework, the slash operators of categorial grammar
must be enriched
with the knowledge about the type of process and the type of morpheme.
We adopt a representation similar to Hoeksema and Janda's~\cite{hoeksema88}
notation for
the operator. The 3-tuple ({\em direction, morpheme type, process type})
indicates direction\footnote{
we have not yet incorporated into our model the word-order
variation in syntax. See~\cite{hoffman92} for a CCG based approach
to this phenomenon.} (left, right, unspecified), morpheme type
(free, bound), and the type of morphological or syntactic
attachment (e.g., affixation, syntactic concatenation,
reduplication\footnote{
intensifiers such as {\exm ap-} and {\exm bes-} in
{\exm ap-a\c{c}{\i}k} and {\exm bes-belli} may appear as prefixes but they
are in fact reduplicated from the first syllable of the stem}, clitic).
Examples
of different operator combinations are as follows:\vspace*{4mm}

%\begin{table}
{\small
\begin{tabular}{lll}
Operator & Morpheme & Example\hspace*{2cm}\\ & & \\
$<\setminus$, bound, clitic$>$ & {\exm de} &
\begin{minipage}[t]{7cm}\small\begin{tabular}{lll}
              {\exm Ben} & {\exm de} & {\exm yaz-ar-{\i}m}\\
              I & too & write-TENSE-PERS
           \end{tabular}\par
           'I write too.'
\end{minipage}\\ & & \\
$<\setminus$, bound, affix$>$ & {\exm -de} &
\begin{minipage}[t]{7cm}\small\begin{tabular}{lll}
              {\exm Ben-de} & {\exm kalem} & {\exm var}\\
              I-LOCATIVE & pen  & exist
           \end{tabular}\par
           'I have a pen.'
\end{minipage}\\ & & \\
$</$, bound, redup$>$ & {\exm ap-} &
\begin{minipage}[t]{7cm}\small\begin{tabular}{ll}
              {\exm ap-a\c{c}{\i}k} & {\exm durum} \\
              INT-clear & situation
           \end{tabular}\par
           'Very clear situation'
\end{minipage}\\ & & \\
$</$, free, concat$>$ & {\exm uzun} &
\begin{minipage}[t]{7cm}\small\begin{tabular}{ll}
              {\exm uzun} & {\exm yol} \\
              long & road
           \end{tabular}\par
           'long road'
\end{minipage}\\ & & \\
$<\setminus$, free, concat$>$ & {\exm ba\c{s}ka} &
\begin{minipage}[t]{7cm}\small\begin{tabular}{ll}
              {\exm bu-ndan} & {\exm ba\c{s}ka} \\
              this-ABLATIVE & other
           \end{tabular}\par
           'other than this'
\end{minipage}\\ & & \\
$<\mid$, free, concat$>$ & {\exm oku} &
\begin{minipage}[t]{7cm}\small\begin{tabular}{lll}
              {\exm adam} & {\exm kitab-{\i}} & {\exm oku-du} \\
              man & book-ACC & read-TENSE
           \end{tabular}\\
           or \\
           {\exm adam okudu kitab{\i}}\\
           'The man read the book'
\end{minipage}\\  & & \\
\end{tabular}
}
%\caption{Operators in the proposed model.}
%\label{optable}
%\end{table}

\section{Information Structure and Tactical Constraints}
\label{fssect}
Entries in the categorial lexicon have tactical constraints, grammatical
and semantic features, and phonological representation. Similar
to {\sc HPSG}~\cite{pollard94}, every entry
is a signed attribute-value matrix.

Syntactic and semantic information are of grammatical ($g$) sign and
semantic ($s$) sign, respectively. These properties include
agreement features such as person, number, and possessive, and
selectional restrictions:

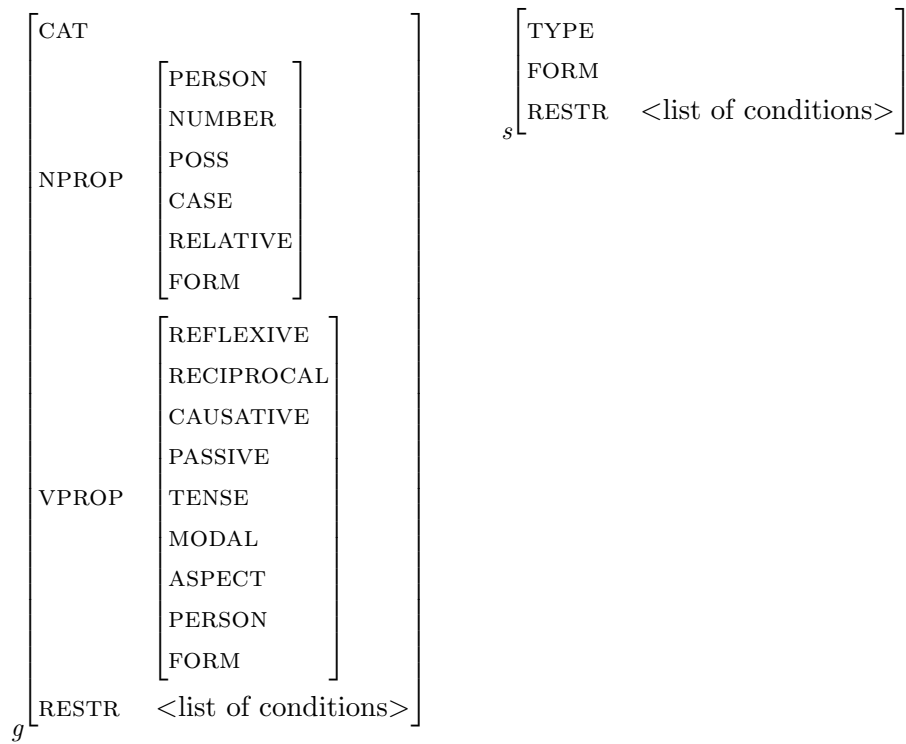
\avmsp
\begin{avm}
\sort{g}{\[ cat \\ nprop &
                  {\[ person \\ number \\ poss \\ case \\ relative \\ form
                  \]} \\
            vprop & {\[ reflexive \\ reciprocal \\ causative \\ passive \\
                      tense \\ modal \\ aspect \\ person \\ form\]}\\
            restr & $<$list of conditions$>$
        \]}
\end{avm}\hspace{5mm}\hspace*{4mm}
\avmsp
\begin{avm}
\sort{s}{\[ type \\ form \\
        restr & $<$list of conditions$>$
        \]}
\end{avm}

Basic and derived categories of CG are of $p$ (property) or $f$
(function) sign, respectively.

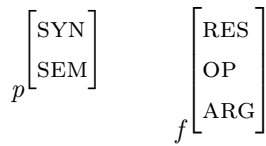
\avmsp
\begin{avm}
\sort{p}{\[ syn \\ sem \]}
\end{avm}\hspace{5mm}\hspace*{4mm}
\begin{avm}
\sort{f}{\[ res \\ op \\ arg \]}
\end{avm}
\avmsp

{\attr res-op-arg} is the categorial notation for the element. Every
{\attr res} and {\attr arg} feature has an $f$ or $p$ sign.

Lexical and phrasal elements have functional representation ($f$ or $p$ sign)
and the {\attr phon} feature.
{\attr phon} represents the phonological string. Lexical elements
may have (a) phonemes, (b) meta-phonemes
such as {\sc H} for high vowel, and {\sc D} for a dental stop whose
voicing is not yet determined, and (c) optional segments,
e.g., {\exm -(y)lA}, to model vowel/consonant drops,  in the {\attr phon}
feature. During composition,
the surface forms of composed elements are mapped and saved in {\attr phon}.
{\attr phon} also allows efficient lexicon search. For instance, the causative
suffix {\exm -DHr} has eight different realizations but only one
lexical entry.

A special feature value called {\tt none} is used for imposing certain
morphotactic constraints. For instance, most of the inflectional
morphemes of Turkish have the category {\sc X}$\setminus${\sc X} where
{\sc X} is the category of the stem. {\tt none} is used to make sure
that the stem is not inflected with the same feature more than once; it
also ensures, through {\attr syn} constraints, that inflections are
marked in the right order.
A sample lexicon entry for a derivational suffix is given in
Figure~\ref{lu}. For composition, we use a generalized LR
parser~\cite{tomita87}
in which CCG rules are encoded as recursive rewrite rules with
equational constraints.

\begin{figure}[h]
\begin{center}

{\small
\newbox\b
\newbox\c

\setbox\b=\hbox{\begin{avm}
\[{} syn & \[{} cat & $n$ \\
                  nprop & \[{} person & {\tt none} \\
                               number & {\tt none}\\
                               possessive & {\tt none} \\
                               case & {\tt none} \\
                               relative & {\tt none} \\
                               form & common \] \] \\
      sem & \[{} type & property \\
              form & has(\@2, \@1) \]
\]
\end{avm}}

\setbox\c=\hbox{\begin{avm}
\[{} syn & \[{} cat & $n$ \\
              nprop & \[{} form & common or proper \] \] \\
      sem & \[{} type & entity \\
              form & \@2 \]
\]
\end{avm}}

\begin{avm}
\[{} res & \[{} res & \box\b  \\
                  op & (/, free, concat) \\
                  arg & \box\c \] \\
  op & ($\setminus$, bound, affix) \\
  arg & \[{} syn & \[{} cat & $n$ \\
                      nprop & \[{} person & {\tt none} \\
                                    number & singular\\
                                    possessive & {\tt none} \\
                                    case & {\tt none} \\
                                    relative & {\tt none} \\
                                    form & common \] \] \\
              sem & \[{} type & entity \\
                      form & \@1 \] \] \\
  phon & "lH"\]
\end{avm}

}  %fontsize

\end{center}
\caption{Lexicon entry for {\exm -lH}.}
\label{lu}
\end{figure}

\section{Conclusion}

Turkish is a language in which grammatical functions can be marked
morphologically (e.g., case), or syntactically (e.g., indirect
objects). Semantic composition is also affected by the interplay
of morphology and syntax, for instance the change in the scope
of modifiers and genitive suffixes, or valency and thematic
role change in causatives. To model interactions between domains,
we propose a categorial approach in which composition in all domains
proceed in parallel.
In the domain of phonology, there are categorial accounts of
prosody \cite{steedman91} and voice assimilation \cite{wheeler88}.
Our treatment of phonology is not yet integrated into the uniform
grammar architecture. Morphophonemic processes such as vowel
harmony and devoicing are modeled as mappings from the operator
and the phonological strings to surface forms. Integrating categorial
phonology into the architecture will help restore the modularity of
processing at all domains.

\end{document}